\documentclass[times, review, 10pt]{elsarticle}
\usepackage[%
  top=3.3cm,      
  right=3.8cm,    
  bottom=3.3cm,   
  left=3.8cm      
]{geometry}
\usepackage{setspace}
\usepackage{amssymb}
\usepackage{lipsum}
\usepackage{float}
\usepackage{hyperref}

\journal{Astronomy $\&$ Computing}

\begin{document}

\begin{frontmatter}


\makeatletter
\renewcommand{\@maketitle}{%
  \begin{center}%
    {\fontsize{14pt}{16pt}\selectfont\@title\par}%
    \vskip 1em%
    {\fontsize{8pt}{10pt}\selectfont
     \lineskip .5em%
     \begin{tabular}[t]{c}\@author\end{tabular}\par}%
    \vskip 1em%
  \end{center}%
}
\makeatother

\title{Multimodal Connectome Fusion via Cross-Attention for Autism Spectrum Disorder Classification Using Graph Learning}

\author[a,b]{Ansar Rahman\thanks{Corresponding author: ansar@example.com}}
\author[c]{Hassan Shojaee-Mend}
\author[a,b,d]{Sepideh Hatamikia}

\affiliation[a]{organization={Department of Medicine, Danube Private University (DPU)},
                city={Krems},
                country={Austria}}

\affiliation[b]{organization={Department of Medical Physics and Biomedical Engineering, Medical University of Vienna},
                city={Vienna},
                country={Austria}}

\affiliation[c]{organization={Department of Medical Informatics, Faculty of Medicine, Gonabad University of Medical Sciences},
                city={Gonabad},
                country={Iran}}

\affiliation[d]{organization={Austrian Center for Medical Innovation and Technology (ACMIT)},
                city={Wiener Neustadt},
                country={Austria}}

\begin{abstract}
Autism spectrum disorder (ASD) is a complex neurodevelopmental condition characterized by atypical functional brain connectivity and subtle structural alterations. Resting-state fMRI (rs-fMRI) has been widely used to identify disruptions in large-scale brain networks, while structural MRI (sMRI) provides complementary information about morphological organization. Despite their complementary nature, effectively integrating these heterogeneous imaging modalities within a unified framework remains challenging. This study proposes a multimodal graph learning framework that preserves the dominant role of functional connectivity while integrating structural imaging and phenotypic information for ASD classification.

The proposed multimodal graph-based framework is evaluated on the multi-site ABIDE-I dataset. Each subject is represented as a node within a population graph. Functional and structural features are extracted as modality-specific node attributes, while inter-subject relationships are modeled using a pairwise association encoder (PAE) based on phenotypic information. Two edge-variational graph convolutional networks (EV-GCNs) are trained to learn subject-level embeddings. To enable effective multimodal integration, we introduce a novel asymmetric transformer-based cross-attention mechanism that allows functional embeddings to selectively incorporate complementary structural information while preserving functional dominance. The fused embeddings are then passed to a feed-forward multilayer perceptron (MLP) for ASD classification.

Using stratified 10-fold cross-validation, the framework achieved an AUC of 87.3\% and an accuracy of 84.4\%. Under leave-one-site-out cross-validation (LOSO-CV), the model achieved an average cross-site accuracy of 82.0\%, outperforming existing methods by approximately 3\% under 10-fold cross-validation and 7\% under LOSO-CV. The proposed framework effectively integrates heterogeneous multimodal data from the multi-site ABIDE-I dataset, improving automated ASD classification across imaging sites.

\end{abstract}




\end{frontmatter}




\section{Introduction}
\label{introduction}

The complex neurodevelopmental disorder known as ASD is characterized by limited and repetitive patterns of behavior, as well as ongoing impairments in social interaction and communication \citep{lord2020autism}. ASD manifests clinically in a very diverse way, with significant individual differences in intensity of symptoms, developmental pathways, and underlying neurobiological processes. Because of this heterogeneity, objective diagnosis is severely hampered, and there is increasing interest in data-driven methods that can identify diffuse and subtle neural patterns.

Neuroimaging research has increasingly shown that ASD is linked to widespread changes in both brain structure and functional connectivity rather than isolated abnormalities in particular brain regions \citep{ecker2015neuroimaging}. Anatomical organization and morphological characteristics can be thoroughly understood through sMRI, whereas rs-fMRI records spontaneous neural activity and functional interactions throughout dispersed brain networks. No single imaging modality can adequately capture the intricate neurobiological signatures of ASD, even though each one provides insightful information. Thus, one of the main challenges in ASD research continues to be the efficient integration of complementary multimodal information.

For the classification and characterization of ASD, automated analysis of high-dimensional neuroimaging data has become more popular due to advancements in machine learning, especially deep learning \citep{dcouto2024multimodal, qiang2023deep}. Single-modality functional connectivity was the main focus of early deep learning techniques. For instance, the efficacy of transfer learning on extensive ASD datasets and using autoencoders to learn latent representations of whole-brain functional connectivity was shown \citep{heinsfeld2018identification,sherkatghanad2020automated}. In order to supplement imaging data and enhance diagnostic performance, researchers \citep{quaak2021deep} started adding non-imaging variables within a machine learning framework, such as age, sex, genetic information, and cognitive measures, as larger cohorts and richer non-imaging phenotypic data became available. However, the representational capacity of conventional machine learning techniques is frequently limited by the high dimensionality and heterogeneous nature of such multimodal data \citep{zhang2020survey}.

Thus, deep-learning-based multimodal fusion techniques have attracted interest for their ability to jointly model heterogeneous data sources. Previous research has demonstrated the efficacy of deep architectures in related neuroimaging tasks, including sex prediction and brain age estimation \citep{peng2021accurate}.  Using large, diverse datasets, convolutional neural networks (CNNs) have also been investigated for the classification of ASD \citep{khosla2019ensemble}. Regardless of their achievements, these non-graph-based methods are frequently limited to single-modality inputs and usually work with Euclidean data representations, which limits their capacity to model population-level structure and inter-subject relationships explicitly. At the same time, although multimodal learning has attracted increasing attention in ASD research, most existing approaches remain limited in how multiple data sources are integrated. In practice, many studies primarily combine rs-fMRI with non-imaging or demographic variables such as age, sex, and acquisition site. While these variables provide important contextual information, they do not fully exploit the complementary neurobiological insights available from sMRI. In particular, sMRI captures morphological and anatomical variations that are not directly observable in functional connectivity alone. To date, multimodal integration of rs-fMRI and sMRI remains a rarely explored area in ASD classification.

To address the limitations of conventional multimodal approaches, graph-based learning frameworks have emerged as a promising alternative. Unlike standard deep learning models that process subjects independently, graph models explicitly encode relationships between individuals at the population level. By representing subjects as nodes and modeling inter-subject relationships through edges, graph neural networks enable the joint use of imaging and non-imaging information within a unified representation \citep{parisot2018disease}. Graph convolutional networks (GCNs) further extend convolutional operations to non-Euclidean domains, making them well-suited for capturing subject associations and brain connectivity patterns \citep{bianchi2021graph}. Early work demonstrated that population graphs constructed using non-imaging phenotypic similarities could improve ASD classification performance \citep{parisot2018disease}, motivating a growing body of graph-based ASD research. Subsequent studies have explored increasingly sophisticated graph architectures, including hierarchical GCNs that jointly model subject-level associations and network topology \citep{jiang2020hi}, as well as adaptive graph constructions such as the Edge-Variational GCN (EV-GCN) by \citep{huang2020edge}, which learns edge weights through a pairwise association encoder. While these methods have shown promising results, most graph-based ASD studies remain limited in two important aspects. First, they primarily rely on a single imaging modality, typically rs-fMRI augmented with non-imaging phenotypic information. Second, even when deeper GCN architectures are employed \citep{cao2021using}, increasing network depth alone does not address the challenge of effectively integrating complementary structural and functional information across different modalities. Therefore, despite these advances, multimodal graph-based ASD classification remains relatively underexplored, particularly with respect to joint rs-fMRI and sMRI integration. To the best of our knowledge, only a limited number of studies have explicitly incorporated both structural MRI and functional MRI alongside non-imaging phenotypic data within the ABIDE dataset. Dong et al. \citep{dong2025framework} conducted a comprehensive and standardized comparison of multiple machine learning models using functional connectivity, structural volumetric measures, and non-imaging phenotypic features. However, this study primarily relies on machine-learning-based feature-level integration and ensemble strategies, and also does not explicitly modeled cross-modal interactions or joint learning population graph structure in an end-to-end multimodal framework. Similarly, Ashrafi et al. \citep{ashrafi2025enhanced} proposed a multi-branch GCN architecture that integrates rs-fMRI, sMRI, and non-imaging phenotypic information from the ABIDE dataset, demonstrating improved classification performance over several baselines. While effective, the proposed approach processes each modality largely independently before fusion through concatenation, which limits explicit modeling of cross-modal dependencies and constrains joint optimization of multimodal representations.

Advances in multimodal learning have highlighted the importance of modeling interactions between heterogeneous data sources rather than relying on simple feature concatenation. Fixed fusion strategies often fail to capture nuanced, context-dependent relationships between modalities, particularly when functional and structural brain representations encode distinct but interrelated aspects of neurobiology. Recently, attention-based mechanisms have emerged as an effective approach for selectively integrating complementary information across modalities by explicitly modeling inter-modality dependencies \citep{xu2023multimodal}. Within a graph learning context, such mechanisms offer the potential to align modality-specific node representations while preserving population-level relational structure.

\begin{figure}[H]
    \centering
    \includegraphics[
        width=\columnwidth,
        trim=275 0 270 0,
        clip
    ]{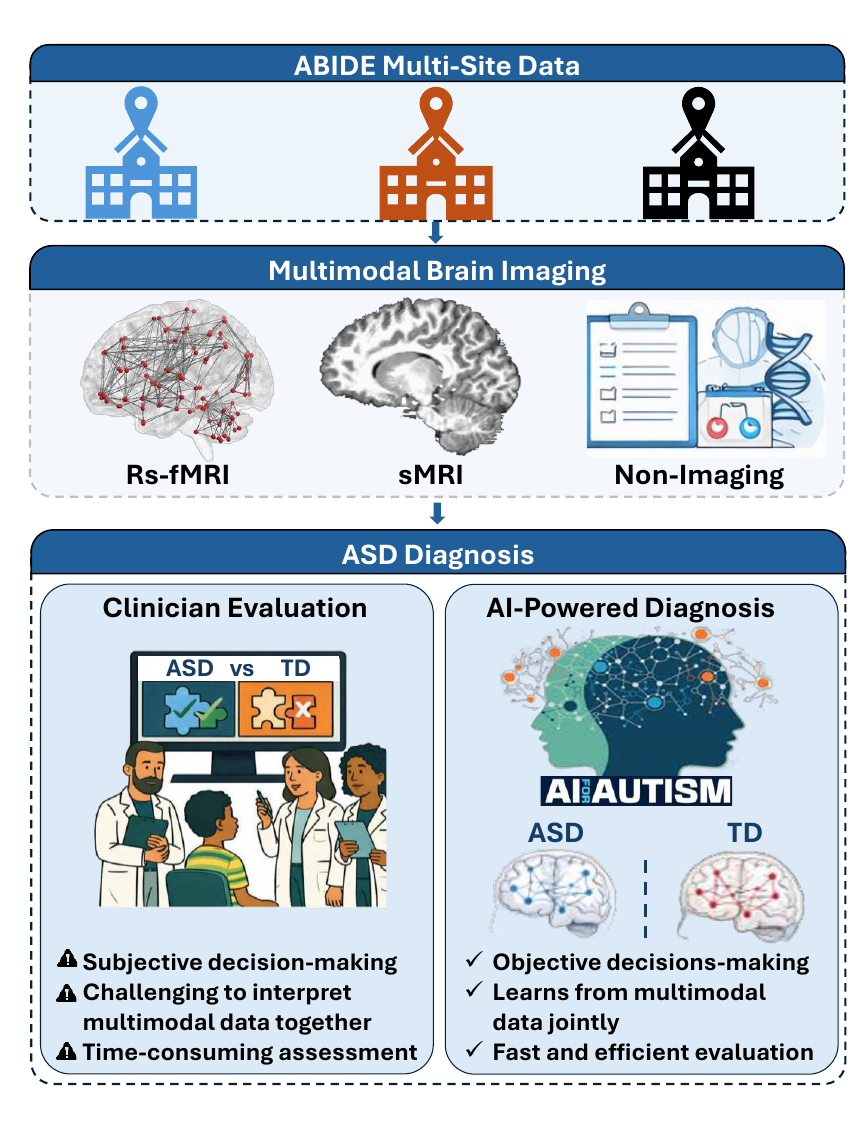}
    \caption{Overview of multimodal ASD diagnosis using heterogeneous ABIDE multi-site data, highlighting the transition from clinician-based assessment to AI-driven multimodal analysis}
    \label{Problem_definition}
\end{figure}

To address the existing challenges, we propose a multimodal graph learning framework for ASD classification that jointly integrates functional and structural brain connectivity derived from rs-fMRI and sMRI using an asymmetric cross-attention-based fusion mechanism. By treating modality-specific population graphs as complementary representations of brain organization, the proposed approach explicitly learns how information from one modality can guide and refine feature representations in the other. This design enables adaptive weighting of modality-specific features, facilitates the extraction of discriminative connectivity patterns, and enhances robustness to site-related variability. To evaluate the proposed framework, we conduct extensive experiments on the large and heterogeneous multi-site ABIDE-I neuroimaging dataset, comprising data collected from 17 acquisition centers (CALTECH, CMU, KKI, LEUVEN, MAX-MUN, NYU, OHSU, OLIN, PITT, SBL, SDSU, STANFORD, TRINITY, UCLA, UM, USM, and YALE). The dataset includes multimodal brain imaging and phenotypic information acquired across diverse clinical sites, reflecting the heterogeneity commonly encountered in real-world ASD studies. In clinical practice, ASD diagnosis primarily relies on behavioral observations and interview-based assessments as illustrated in Fig.~\ref{Problem_definition}, which can be subjective and time-consuming \citep{song2021machine}. In contrast, recent advances in machine learning demonstrate that AI-based analysis of neuroimaging data can provide more objective and efficient diagnostic support by automatically learning discriminative patterns from multimodal brain data \citep{santana2022rs,pang2026multi}. Experimental results show that the proposed framework effectively leverages multimodal information and consistently outperforms existing state-of-the-art graph-based and multimodal fusion approaches for ASD classification.

The main contributions of the proposed work are summarized as follows:

\begin{itemize}

\item We propose a multimodal population graph learning framework for ASD classification that models rs-fMRI and sMRI connectivity using separate modality-specific population graphs with shared phenotypic-driven edge weights, enabling robust integration of structural and functional information across sites. 

\item We introduce a novel transformer-based asymmetric cross-attention–based fusion mechanism that captures directional inter-modality dependencies by using rs-fMRI features as queries and sMRI features as keys and values, enabling adaptive integration of complementary structural information while preserving the dominant role of functional connectivity. 

\item We perform comprehensive evaluations on the multi-site ABIDE-I dataset using both stratified 10-fold cross-validation and LOSO-CV to rigorously assess classification performance and cross-site generalization. Compared to existing state-of-the-art multimodal and graph-based approaches, the proposed method demonstrates performance improvements of approximately 3\% (in 10-fold cross-validation) and 7\% (in LOSO-CV), highlighting enhanced discriminative capability, improved robustness, and superior generalization across imaging sites. 

\end{itemize}

\section{Proposed Methodology}

In this study, we propose an end-to-end population-graph–based multimodal learning framework for ASD classification that integrates rs-fMRI, sMRI, and non-imaging phenotypic information. Each subject is represented as a node in modality-specific population graphs, where node attributes are derived from imaging-based connectivity features and inter-subject relationships are modeled through adaptive, learnable edge weights computed from non-imaging attributes. To this end, we used a Pairwise Association Encoder (PAE) that constructs a data-driven population graph by learning non-imaging phenotypic affinities in a latent space optimized for the downstream classification task.

\begin{figure}[H]
    \centering
    \includegraphics[width=\textwidth, trim=80 42 80 35, clip]{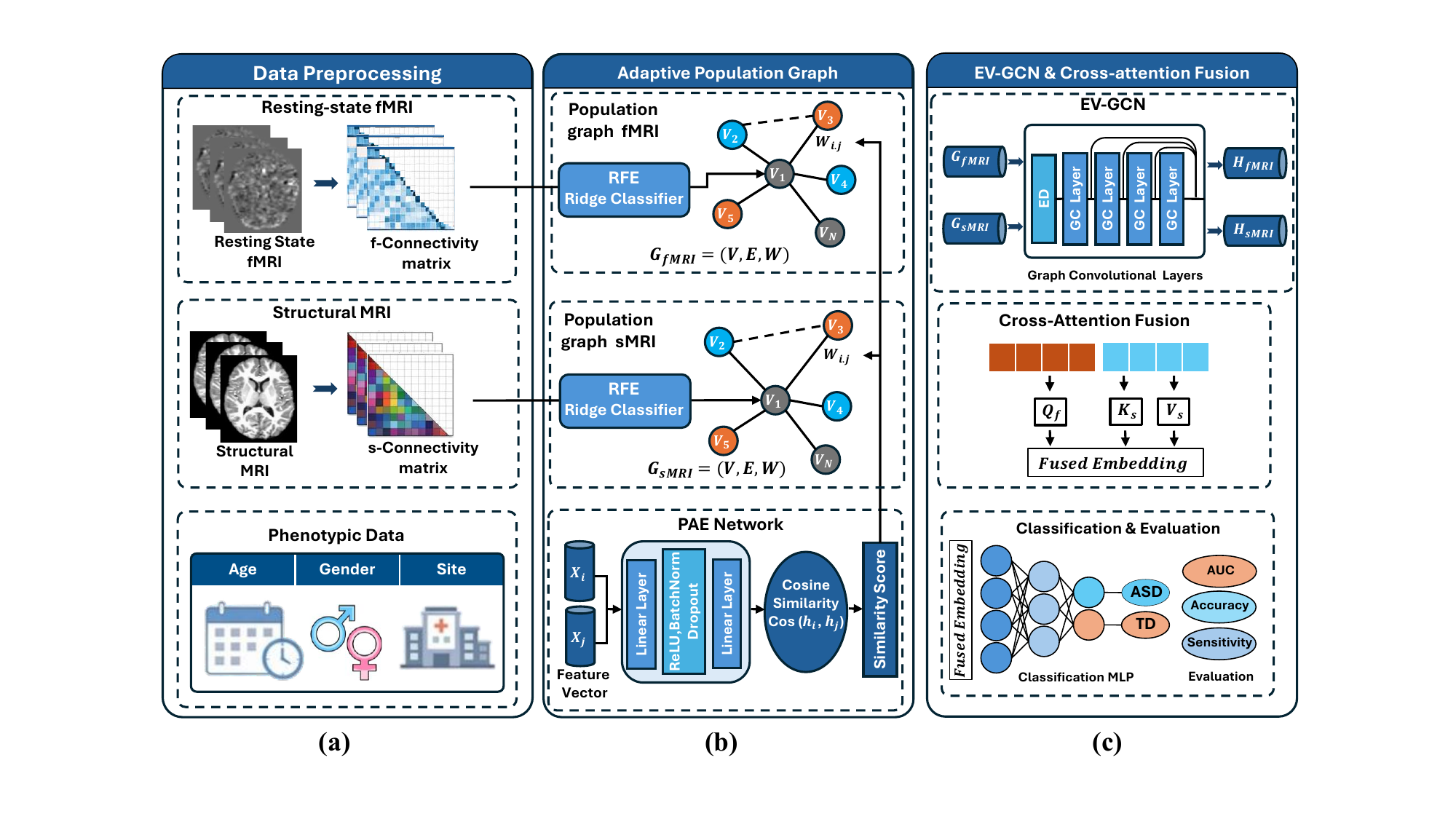}
    \caption{End-to-End Workflow Diagram(a) rs-fMRI, sMRI, and phenotypic data are processed to extract functional and structural connectivity features; (b) modality-specific population graphs are constructed and processed using EV-GCN to learn functional and structural node embeddings;(c) the resulting embeddings are integrated through asymmetric cross-attention fusion and passed to a Feed Forward MLP classifier to predict ASD vs. TD}
    \label{fig_flowdiagram}
\end{figure}

Modality-specific EV-GCNs \citep{huang2020edge} are employed to learn complementary subject-level representations from rs-fMRI and sMRI population graphs, jointly optimizing node embeddings and graph structure in an end-to-end manner. The resulting modality-specific embeddings are then integrated using an asymmetric cross-attention fusion mechanism, in which rs-fMRI representations act as queries, and sMRI representations serve as keys and values. This design explicitly prioritizes functional connectivity while allowing structural information to selectively modulate functional representations. Finally, the fused embeddings are passed to a feed-forward MLP for binary ASD classification. An overview of the complete framework is illustrated in Fig. \ref{fig_flowdiagram}.

\subsection{Dataset and Preprocessing}

The proposed framework was evaluated on the ABIDE-I dataset \citep{di2014autism}, which provides rs-fMRI, sMRI, and non-imaging phenotypic data for ASD and typically developing (TD) subjects. Following the curation protocol in \citep{abraham2017deriving} to ensure fair comparison with prior population-graph approaches \citep{abraham2017deriving,kazi2019inceptiongcn}, subjects with incomplete coverage or inconsistent scanner information were excluded, yielding 871 participants (403 TD, 468 ASD).

Experiments were conducted using both stratified 10-fold and LOSO-CV. In each fold, feature selection and population graph construction were performed exclusively on the training set to prevent information leakage. Test subjects’ imaging and non-imaging phenotypic information were not used during training.

\textbf{rs-fMRI:}
Preprocessed rs-fMRI data were obtained from the ABIDE Preprocessed Connectomes Project \citep{craddock2013imaging}. Regional mean BOLD time series were extracted via atlas-based parcellation, and Pearson correlation was used to compute symmetric functional connectivity matrices as shown in Fig.\ref{fig_flowdiagram} (a). Brain regions were defined using the Harvard–Oxford (HO) atlas (111 ROIs), whose probabilistic definition accommodates inter-subject anatomical variability in multi-site data.

\textbf{sMRI:}
sMRI data were processed using FreeSurfer \citep{fischl2012freesurfer} with the standard \textit{recon-all} pipeline for automated segmentation of cortical, subcortical, and white matter regions. Each brain was parcellated using the AAL atlas (116 ROIs), and region-wise morphological connectivity matrices were computed to capture inter-regional structural relationships. Only the upper-triangular values of these symmetric matrices were extracted as node features for each subject, as shown in Fig.\ref{fig_flowdiagram} (a). This representation encodes morphological variations across cortical thickness, subcortical volumes, and white matter properties while maintaining a compact feature set.

\textbf{Non-Imaging Phenotypic Data:}
The most commonly used non-imaging features in literature \citep{dong2025framework,huang2020edge,wang2023multimodal} Age, gender, and acquisition site were selected to capture inter-subject similarity in the multi-site setting. These variables were numerically encoded and used solely for population graph construction through the Pairwise Association Encoder (PAE) as depicted in Fig.\ref{fig_flowdiagram} (b), which learns adaptive edge weights. They were not used as node features or direct inputs to the classifier, thereby avoiding information leakage while incorporating non-imaging phenotypic context into graph learning.

\subsection{Multimodal Brain Network Construction}

In this study, multimodal brain networks are constructed from standardized, preprocessed connectivity features derived from the ABIDE-I dataset, as described in Section 2.1. Rather than operating on raw neuroimaging volumes, we focus on population-level graph modeling \citep{huang2020edge}, which improves reproducibility across acquisition sites and enables effective integration of multimodal information.

Each subject is represented as a node in a population graph, with node attributes derived from modality-specific connectivity features. For rs-fMRI and sMRI modalities based on connectome data, distinct graphs are constructed as shown in Fig.\ref{fig_flowdiagram} (b), allowing each modality to maintain its inherent representational properties while sharing a connectivity structure driven by non-imaging phenotypic data. For each modality, vectorized connectivity features are used as node attributes, while inter-subject similarities are encoded through adaptive edge weights derived from non-imaging phenotypic information. These population graphs serve as the basis for graph convolutional representation learning.

\subsubsection{Feature Selection and Standardization}
Connectivity matrices derived from both resting-state functional and structural imaging are inherently high-dimensional due to the large number of inter-regional connections. For each modality, the symmetric connectivity matrix $C \in \mathbb{R}^{R \times R}$ was transformed into a feature vector by extracting the upper triangular elements while excluding the diagonal, where $R$ represents the number of brain regions. This procedure retains all unique pairwise relationships and removes redundant information arising from matrix symmetry.

To reduce the risk of overfitting, feature selection was performed separately for each modality within every training fold. Recursive Feature Elimination (RFE) \citep{ding2022efficient}  with ridge regression was used to rank features based on their contribution to classification performance. At each iteration, features with the smallest weight magnitudes were removed, and a fixed number of features ($D = 2400$) was retained for each modality. The selected features were subsequently standardized using z-score normalization and used as node attributes in the corresponding modality-specific population graphs. 

\subsubsection{Adaptive Population Graph Construction}

Let $\mathcal{G} = (\mathcal{V}, \mathcal{E}, \mathcal{W})$ denote a population graph, where $\mathcal{V}$ represents a set of $N$ nodes corresponding to
subjects, $\mathcal{E}$ is the set of edges, and $\mathcal{W}$ contains edge weights. Each node $v_i \in \mathcal{V}$ is associated with an imaging-derived feature vector $\mathbf{z}_i \in \mathbb{R}^{D}$, which captures discriminative diagnostic information from imaging modalities such as rs-fMRI and sMRI.

To model inter-subject relationships, we define graph connectivity using non-imaging attributes (sex, age, site), which provide complementary contextual information. Instead of relying on fixed or statistically defined similarities, we used a learnable function to model edge weights. Specifically, the edge weight $w_{ij} \in \mathcal{W}$ between  subjects $i$ and $j$ is computed as
\begin{equation}
w_{ij} = f(\mathbf{x}_i, \mathbf{x}_j; \Omega),
\end{equation}
where $\mathbf{x}_i$ and $\mathbf{x}_j$ denote non-imaging features and $\Omega$ represents trainable parameters of the \textit{Pairwise Association Encoder (PAE)}.

\subsubsection{Pairwise Association Encoder (PAE)}
The PAE first normalizes each modality of the non-imaging features to zero mean and unit variance to reduce scale discrepancies across modalities \citep{parisot2018disease}. The normalized features are then projected into a shared latent space via a multi-layer perceptron (MLP). In this work, the latent dimensionality is set to $D_h = 128$.

The association strength between subjects is computed using cosine similarity in the latent space:
\begin{equation}
w_{ij} = \frac{\mathbf{h}_i^{\top}\mathbf{h}_j}{2\|\mathbf{h}_i\|\|\mathbf{h}_j\|} + 0.5,
\end{equation}
which rescales similarity scores to ensure numerical stability. Defining edge weights in the latent space yields more robust graph connectivity than directly operating on raw non-imaging features.

\subsubsection{Edge-Variational Graph Convolution on Adaptive Population Graphs}

Given the adaptively constructed population graph $\mathcal{G} = (\mathcal{V}, \mathcal{E}, \mathcal{W})$,  EV-GCN performs edge-weighted graph convolution to propagate information across subjects and learn population-aware node representations. Each node corresponds to a subject and is initialized with modality-specific connectivity features, while edges encode adaptive inter-subject relationships learned from non-imaging phenotypic information via the Pairwise Association Encoder (PAE).

Graph convolution is performed using spectral filtering based on Chebyshev polynomial approximation, which enables efficient aggregation of information from multi-hop neighborhoods without explicitly computing eigen-decompositions. In this formulation, learned edge weights directly modulate the message-passing process, controlling the influence of neighboring subjects during feature aggregation.

Since these edge weights are generated by the differentiable PAE, gradients from the classification objective propagate jointly through both the graph convolution layers and the edge-weight learning module. This joint optimization enables simultaneous learning of node embeddings and population graph structure, allowing inter-subject relationships to adapt dynamically to the downstream ASD classification task.

\subsubsection{EV-GCN Encoder}

For each imaging modality, a dedicated EV-GCN encoder \citep{huang2020edge} learns population-aware subject representations from graph-structured connectivity data. Each encoder consists of multiple stacked edge-weighted graph convolution layers with shared hidden dimensionality. Nonlinear activation functions and feature-level dropout are applied after each layer to improve generalization and reduce overfitting. EV-GCN block diagram is depicted in Fig.\ref{fig_flowdiagram} (c).

To preserve information across multiple neighborhood scales and prevent over-smoothing, node embeddings from all graph convolution layers are concatenated using a multi-scale aggregation strategy. This allows the model to integrate both local connectivity patterns and global population-level structures, capturing diagnostically relevant differences across subjects.

The resulting embeddings provide a rich, structured representation for each subject and serve as the input for the subsequent cross-attention-based multimodal fusion module.

\subsection{Transformer-based Asymmetric Cross-Attention for Multimodal Fusion}

\begin{figure}[H]
    \centering
    \includegraphics[
        width=\columnwidth,
        trim=320 120 320 20,
        clip
    ]{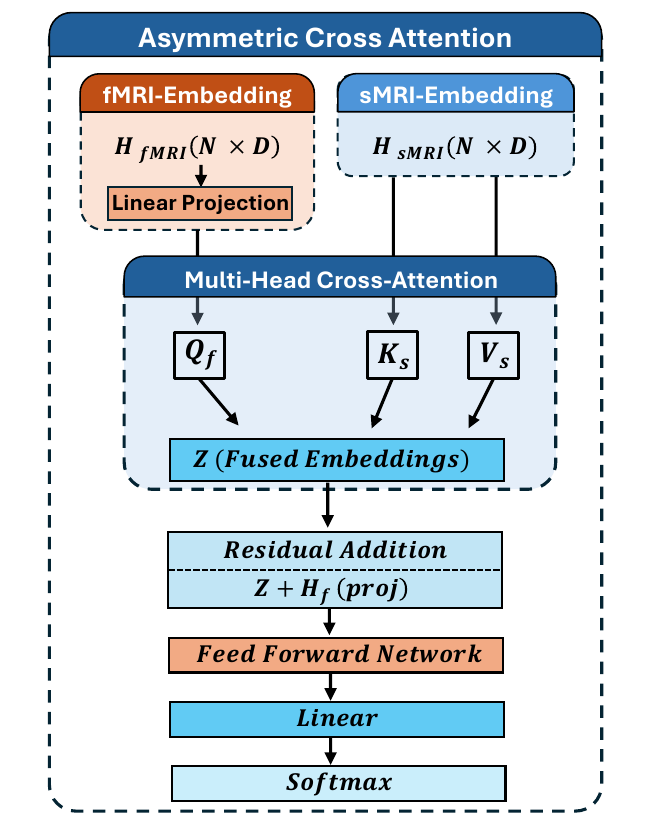}
    \caption{Transformer-based asymmetric cross-attention fusion integrating modality-specific EV-GCN embeddings from functional and structural brain connectivity.}
    \label{Asymmetric Cross-Attention}
\end{figure}

Building upon the modality-specific embeddings learned from the EV-GCN encoders, we proposed a novel transformer-based asymmetric cross-attention fusion mechanism to integrate functional and structural brain connectivity. This mechanism enables rs-fMRI embeddings to selectively attend to complementary information from the sMRI, capturing cross-modal interactions more effectively than conventional feature concatenation or averaging, and forming a key component of our multimodal framework as shown in Fig.\ref{Asymmetric Cross-Attention}.

Let $H_f \in \mathbb{R}^{N \times D}$ and $H_s \in \mathbb{R}^{N \times D}$ denote the node embeddings learned from the rs-fMRI and sMRI population graphs, respectively, where $N$ is the number of subjects and $D$ is the embedding dimensionality produced by the EV-GCN encoders. 

\subsubsection{Asymmetric Cross-Attention Formulation}
To prioritize functional connectivity while incorporating structural context, in this study, we propose the rs-fMRI embeddings to be treated as queries, while the sMRI embeddings serve as keys and values. The functional embeddings are first projected into a shared latent space, while the structural embeddings are directly used as keys and values without an additional projection to retain their original relational structure, enabling them to act as stable contextual references during attention. 

Formally, the attention inputs are defined as
\begin{equation}
Q_{f} = H_{f}W_Q, \quad
K_{s} = H_{s}, \quad
V_{s} = H_{s}
\end{equation}
where  $W_Q \in \mathbb{R}^{D\times D}$ are learnable projection matrices. Multi-head attention is then applied to compute fused representations:
\begin{equation}
Z = \textit{MultiHeadAttn}(Q_{f}, K_{s}, V_{s}),
\end{equation}
allowing each subject’s functional representation to selectively attend to structurally relevant patterns across the population graph. Because the attention mechanism is driven by the functional query representation, the resulting fusion process is naturally guided by functional connectivity patterns, while structural embeddings provide complementary contextual information. This asymmetric formulation ensures that functional connectivity remains the principal source of information in the fused representation while structural connectivity contributes additional context that enhances cross-modal interaction.

\subsubsection{Residual Refinement}
To stabilize training and preserve the discriminative properties of the functional embeddings, a residual connection is applied between the projected functional embeddings and the attention output, maintaining the intended asymmetry of the fusion process:
\begin{equation}
Z_r = Z + Q_{f}.
\end{equation}
The fused representations are subsequently refined using a feed-forward MLP with layer normalization and non-linear activation:
\begin{equation}
\tilde{Z} = \textit{MLP} (Z_r),
\end{equation}
where the MLP consists of a sequence of layer normalization, linear transformation, \textit{GELU} activation, dropout, and a final linear projection. This refinement step enhances the expressive capacity of the fused embeddings while maintaining the original functional signal.

\subsection{Classification Module}
The refined embeddings $\tilde{Z} \in \mathbb{R}^{N \times D}$ are then passed to a linear classification layer:
\begin{equation}
\hat{Y} = \tilde{Z} W_c + b,
\end{equation}
where $W_c \in \mathbb{R}^{D \times C}$ and $b \in \mathbb{R}^{C}$ are learnable parameters, and $C$ denotes the number of diagnostic classes. The entire framework, comprising the dual EV-GCN encoders, the asymmetric cross-attention fusion module, and the classification layer, is trained end-to-end using the cross-entropy loss:
\begin{equation}
\mathcal{L} = - \sum_{i=1}^{N} \sum_{c=1}^{C} y_{ic} \log \hat{y}_{ic}.
\end{equation}
This joint optimization enables coordinated learning of modality-specific embeddings and cross-modal interactions, effectively capturing complex relationships in functional and structural connectivity for ASD classification.

\section{Experiments and Results}

This section presents a comprehensive evaluation of the proposed multimodal population-graph framework for ASD classification using the ABIDE-I dataset. The experimental analysis investigates the effectiveness of the proposed framework under different modality configurations and examines the contribution of the proposed asymmetric cross-attention–based fusion strategy. 

Experiments are performed using both stratified 10-fold and LOSO cross-validation methods to evaluate robustness and generalization across the sites. The stratified 10-fold cross-validation setup assesses subject-level generalization by preserving class balance across folds, while LOSO-CV specifically examines the model's capacity to generalize to unseen acquisition sites. These assessment techniques provide a comprehensive evaluation of performance even when multi-center neuroimaging datasets exhibit inter-site variation. 

In our experiments, the spectral graph convolution is implemented using a Chebyshev polynomial approximation of order $K = 3$, enabling localized yet expressive graph filtering, while the network depth is defined by $L_G = 4$ graph convolutional layers to facilitate hierarchical feature aggregation across the population graph. All models are optimized using the Adam optimizer with an initial learning rate of $0.01$ and a weight decay of $5 \times 10^{-5}$ to control model complexity. To further reduce overfitting, a dropout rate of $0.2$ is applied during training. Each model is trained for $300$ epochs. All models are developed on Python with experiments conducted on the open-source PyTorch framework, with graph computations handled via PyTorch Geometric, and experiments run on an NVIDIA RTX 4000 Ada Generation GPU.

\subsection{Performance Evaluation}

The proposed model was evaluated by using Accuracy, Area under the Curve, Sensitivity, Specificity, and F1-score evaluation metrics. Together, these metrics provide a comprehensive assessment of classification performance, particularly in multi-site neuroimaging settings where data heterogeneity must be carefully considered.

Accuracy is defined as:
\begin{equation}
\mathrm{Acc} = \frac{TP + TN}{TP + TN + FP + FN}.
\end{equation}

Sensitivity (True Positive Rate) and Specificity (True Negative Rate) are computed as:
\begin{equation}
\mathrm{Sensitivity} = \frac{TP}{TP + FN},
\qquad
\mathrm{Specificity} = \frac{TN}{TN + FP}.
\end{equation}

The F1-score, which provides a balanced measure of precision and recall, is given by:
\begin{equation}
\mathrm{F1} = 2 \cdot \frac{\mathrm{Precision} \cdot \mathrm{Recall}}
{\mathrm{Precision} + \mathrm{Recall}}.
\end{equation}

The Area Under the Receiver Operating Characteristic Curve (AUC) is computed by plotting the true positive rate against the False Positive Rate.

All metrics are computed exclusively on held-out test subjects in each validation split and reported as averages across cross-validation folds and acquisition sites, ensuring an unbiased and statistically reliable assessment of model performance.

\subsection{Effect of Multimodal Learning}

To analyze the contribution of different imaging modalities and non-imaging phenotypic information, we evaluated several configurations of the proposed framework using stratified 10-fold cross-validation while keeping the baseline EV-GCN architecture unchanged. The results are summarized in Table~\ref{tab:ablation_results}.

\begin{table}[H]
\caption{Effect of Multimodal Learning, evaluated using stratified 10-fold cross-validation}
\centering
\small

\begin{tabular}{l c c c c c c}
\hline
Model & Modality & AUC & Acc & Sensitivity  & Specificity  & F1-Score \\
\hline
         & rs-fMRI & 79.40 & 77.40 & 76.70 & 84.41 & 80.00 \\
Proposed & rs-fMRI* & 84.80 & 82.60 & 83.90 & 85.30& 84.10 \\
 Method& sMRI & 73.90 & 73.30 & 73.00 & 81.00 & 76.40  \\
         & sMRI* & 82.30 & 81.00 & \textbf{86.73}& 76.50 & 80.90 \\
         & Multimodal & \textbf{87.30}& \textbf{84.40}& 85.20& \textbf{87.00}& \textbf{85.70} \\
\hline
\multicolumn{7}{l}{\footnotesize \textit{Note: (*) indicates the inclusion of non-imaging data alongside the specified}}\\
\multicolumn{7}{l}{\footnotesize \textit{imaging modality.}}\\
\end{tabular}
\label{tab:ablation_results}
\end{table}

rs-fMRI by itself already produces competitive results, demonstrating the ability of functional connectivity patterns to discriminate in the categorization of ASD. Incorporating non-imaging phenotypic data regularly enhances performance, suggesting that adaptive modeling of inter-subject connections offers useful context at the population level. 

sMRI alone achieves lower performance compared to rs-fMRI, reflecting its limited sensitivity to functional brain dynamics. However, when structural connectivity is combined with non-imaging phenotypic information, classification performance improves substantially, demonstrating the effectiveness of the adaptive population graph in leveraging complementary non-imaging attributes. Moreover, when compared with relevant state-of-the-art approach for the sMRI modality, the proposed model achieves superior performance, as summarized in Table~\ref{tab:SOTA Comparison for sMRI + Non-Imaging}.

\begin{table}[H]
\centering
\small

\caption{Results comparison for sMRI + non-imaging Data , evaluated using stratified 10-fold cross-validation}
\renewcommand{\arraystretch}{1.2}
\begin{tabular}{l l c c }
\hline
Reference & Model & AUC & Acc \\
\hline
\citep{dong2025framework} & FCN & 69.60 & 66.20  \\

\hline
Proposed Method&& 82.30 & 81.00  \\
\hline
\end{tabular}
\label{tab:SOTA Comparison for sMRI + Non-Imaging}
\end{table}

\begin{figure}[H]
    \centering
    \includegraphics[width=\columnwidth]{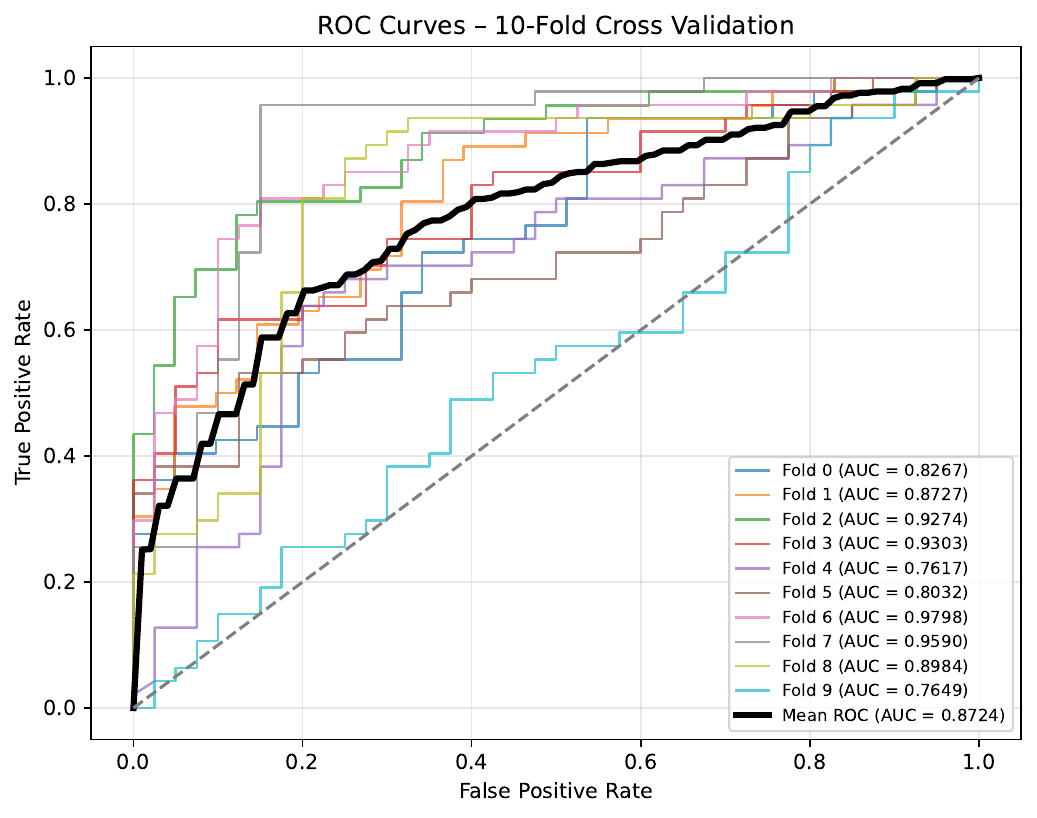}
    \caption{ROC–AUC curve for the proposed model under 10-fold cross-validation}
    \label{fig:roc_auc}
\end{figure}

The best overall performance is achieved through multimodal fusion, where functional and structural information are jointly modeled within the proposed asymmetric cross-attention framework. This configuration attains an AUC of 87.30\%, an accuracy of 84.40\%, a sensitivity of 85.20\%, a specificity of 87.00\%, and an F1-score of 85.70\%, confirming that multimodal integration enables richer and more robust subject representations than any single imaging modality. For results visualization of the proposed model, ROC curves are plotted in Fig.\ref{fig:roc_auc}

\subsection{Effect of asymmetric Cross-Attention Fusion}

To evaluate the effectiveness of the proposed asymmetric cross-attention mechanism for multimodal feature fusion, we compare it against a conventional concatenation \citep{xu2023multimodal} and symmetric cross-attention-based \citep{lu2019vilbert} fusion strategies, using stratified 10-fold cross-validation. In the concatenation approach, modality-specific embeddings are directly merged and passed to the classification head, without explicitly modeling inter-modality interactions, and for symmetric (two-stream) cross attention, the query is exchanged from each modality. In contrast, the proposed asymmetric cross-attention fusion selectively integrates structural information by allowing rs-fMRI to attend sMRI, enabling adaptive and context-aware feature alignment. The quantitative results are reported in Table~\ref{tab:asymmetric_cross_attention_fusion_effect}. Across all assessment measures, the asymmetric cross-attention-based fusion outperforms simple concatenation and symmetric cross-attention. 

\begin{table}[H]
\centering
\small
\caption{Effect of the proposed asymmetric cross-attention–based fusion strategy on multimodal ASD classification performance, evaluated using stratified 10-fold cross-validation }
\begin{tabular}{l c  c  c c c}
\hline
Multimodal Fusion Strategy & AUC & Acc & Sens& Spec & F1-Score  \\
\hline
Concatenation& 84.90 & 83.00 & 84.90& 84.20 & 84.30  \\
Symmetric Cross-Attention & 82.30 & 81.30 & 84.40 & 82.30 & 82.90 \\
Asymmetric Cross-Attention& \textbf{87.30} & \textbf{84.40} & \textbf{85.20} & \textbf{87.00} & \textbf{85.70} \\
\hline
\end{tabular}
\label{tab:asymmetric_cross_attention_fusion_effect}
\end{table}

\subsection{Comparison with State-of-the-Art Methods}

We compare the proposed model with existing population-graph and deep learning–based approaches for ASD classification using stratified 10-fold cross-validation. In particular, the evaluation includes comparisons with several state-of-the-art methods, including Brain Network Transformer \citep{kan2022brain}, Com-BrainTF \citep{bannadabhavi2023community}, ASDFormer \citep{izadi2025asdformer}, DeepGCN \citep{wang2023multimodal}, MMGCN \citep{song2024multi}, GCN \citep{dong2025framework}, EV-GCN \citep{huang2020edge}, and Enhanced EV-GCN \citep{ashrafi2025enhanced}. All these studies used the same 871 number of images from the ABIDE-I dataset within consistent data pre-processing.

\begin{table}[H]
\centering
\small

\caption{Stratified 10-fold cross-validation based results comparison with SOTA ASD classification methods}
\label{tab:sota_results}
\begin{tabular}{l c c c c}
\hline
Reference & Model & Modality & AUC & ACC \\
\hline
\citep{kan2022brain} & Brain Network Transformer & rs-fMRI* & 80.2 & 71.0 \\
\citep{bannadabhavi2023community} & Com-BrainTF & rs-fMRI* & 79.6 & 72.5 \\
\citep{izadi2025asdformer} & ASDFormer & rs-fMRI* & 81.17 & 74.60 \\
\citep{wang2023multimodal} & DeepGCN & rs-fMRI* & 82.59 & 77.27 \\
\citep{song2024multi} & MMGCN & rs-fMRI* & 84.00 & 78.31 \\
\citep{huang2020edge} & EV-GCN & rs-fMRI* & 84.72 & 81.06 \\
\citep{dong2025framework} & GCN & Multimodality & 78.00 & 71.30 \\
\citep{ashrafi2025enhanced} & Enhanced EvGCN & Multimodality & 82.00 & 74.81 \\
\hline
& \textbf{Proposed Method} & \textbf{Multimodality} & \textbf{87.30} & \textbf{84.40} \\
\hline
\multicolumn{5}{l}{\footnotesize \textit{Note:(*) represents integration of non-imaging data into imaging modality}} \\
\multicolumn{5}{l}{\footnotesize \textit{Multimodality represents all modalities, i.e., rs-fMRI, sMRI, and Non-Imaging.}} \\
\end{tabular}
\end{table}

Our proposed multimodal architecture performs better overall across all provided measures, as shown in Table~\ref{tab:sota_results}.  Although current graph learning techniques and transformer-based models enhance performance by modeling population-level interactions and long-range dependencies, they are still limited by their dependence on a single imaging modality \citep{kan2022brain, bannadabhavi2023community, izadi2025asdformer, wang2023multimodal, song2024multi, huang2020edge}.

Among population-graph approaches, EV-GCN \citep{huang2020edge} has shown strong performance by jointly learning node representations and adaptive graph structures from resting-state functional connectivity and non-imaging phenotypic information. Building on this foundation, the proposed framework further enhances classification accuracy and area under the curve by explicitly incorporating structural connectivity alongside functional and non-imaging attributes. Through end-to-end optimization of adaptive edge weights, graph convolutional representations, and a novel asymmetric cross-attention–based fusion mechanism instead of simple concatenation-based fusion like \citep{ashrafi2025enhanced}, helps the model to effectively capture both intra-subject characteristics and inter-subject relationships. This multimodal design enables the extraction of complementary aspects of brain organization that are not fully accessible from functional data alone, resulting in consistent improvements over strong baseline methods and offering a robust solution for multi-site autism classification.

\subsection{Leave-One-Site-Out Based Evaluation}

A LOSO-CV was performed using the ABIDE-I dataset to assess the robustness of the proposed framework to site-specific variability and its capacity to generalize across varied acquisition conditions. Under the LOSO evaluation protocol, data from a single imaging site is held out entirely for testing, while the model is trained on all remaining sites. This process is repeated until each site serves once as the unseen test set, allowing a comprehensive assessment of cross-site generalization. This assessment procedure reveals the model to differences in scanner hardware, acquisition methods, and demographic distributions, offering a rigorous examination of cross-site generalization.

Under LOSO-CV, the proposed approach attains the highest average accuracy of 82.0\%, outperforming existing LOSO-CV style evaluation methods \citep{eslami2019asd}, \citep{almuqhim2021asd}, and \citep{wang2023multimodal} as shown in Table~\ref{tab:site_wise_results}. It achieves superior performance in 12 of the 17 sites as shown in Bar Plots in Fig \ref{fig:roc_auc}, indicating improved robustness to variability across imaging centers. Importantly, the performance gains are evident not only in larger cohorts such as NYU, TRINITY, and UCLA, but also in smaller sites, including CMU, STANFORD, and YALE. This consistent improvement across both large and small datasets suggests that the adaptive population graph construction, together with multimodal fusion, enables the model to better utilize heterogeneous training data and generalize more reliably across different acquisition settings.

\begin{table}[H]
\centering
\small
\caption{Site-wise classification accuracy (\%) under LOSO-CV evaluation}
\begin{tabular}{l c c c c c}
\hline
\textbf{Site} & \textbf{Size} & \textbf{ ASD-} & \textbf{ASD-} & \textbf{DeepGCN} & \textbf{Our } \\
\textbf{ } & \textbf{ } & \textbf{ DiagNet} & \textbf{SAENet} & \textbf{} & \textbf{Proposed} \\
\hline
CALTECH   & 15  & 52.8 & 56.7 & 71.7 & \textbf{86.7} \\
CMU       & 11  & 68.5 & 70.6 & 80.5 & \textbf{81.8} \\
KKI       & 33  & 69.5 & 72.6 & \textbf{76.2} & 63.6 \\
LEUVEN    & 56  & 61.3 & 64.6 & \textbf{74.3} & 71.4 \\
MAX\_MUN  & 46  & 48.6 & 47.5 & 69.7 & \textbf{87.0} \\
NYU       & 172 & 68.0 & 72.0 & 79.1 & \textbf{91.9} \\
OHSU      & 25  & \textbf{82.0} & 72.0 & 78.6 & 72.0 \\
OLIN      & 28  & 65.1 & 66.6 & 80.0 & \textbf{82.1} \\
PITT      & 50  & 67.8 & 73.1 & 72.4 & \textbf{88.0} \\
SBL       & 26  & 51.6 & 56.6 & \textbf{79.7} & 76.9 \\
SDSU      & 27  & 63.0 & 64.2 & 67.7 & \textbf{81.5} \\
STANFORD  & 25  & 64.2 & 53.2 & 68.3 & \textbf{92.0} \\
TRINITY   & 44  & 54.1 & 57.5 & 66.9 & \textbf{95.5} \\
UCLA      & 85  & 73.2 & 68.3 & 72.8 & \textbf{83.5} \\
UM        & 120 & 63.8 & 67.8 & \textbf{81.2} & 78.3 \\
USM       & 67  & 68.2 & 70.0 & 67.5 & \textbf{79.1} \\
YALE      & 41  & 63.6 & 66.0 & 75.6 & \textbf{82.9} \\
\hline
Average   & 51.2 & 63.8 & 64.7 & 74.2 & \textbf{82.0} \\
\hline
\end{tabular}

\label{tab:site_wise_results}
\end{table}

While the model demonstrates consistent improvements across most sites, some performance variability across sites is evident, particularly for smaller cohorts such as KKI, OHSU, and SBL. This trend aligns with previous LOSO studies on ABIDE and is likely due to limited sample sizes and site-specific distribution shifts. 

\begin{figure}[H]
    \centering
    \includegraphics[width=\columnwidth, trim=138 13 132 163, clip]{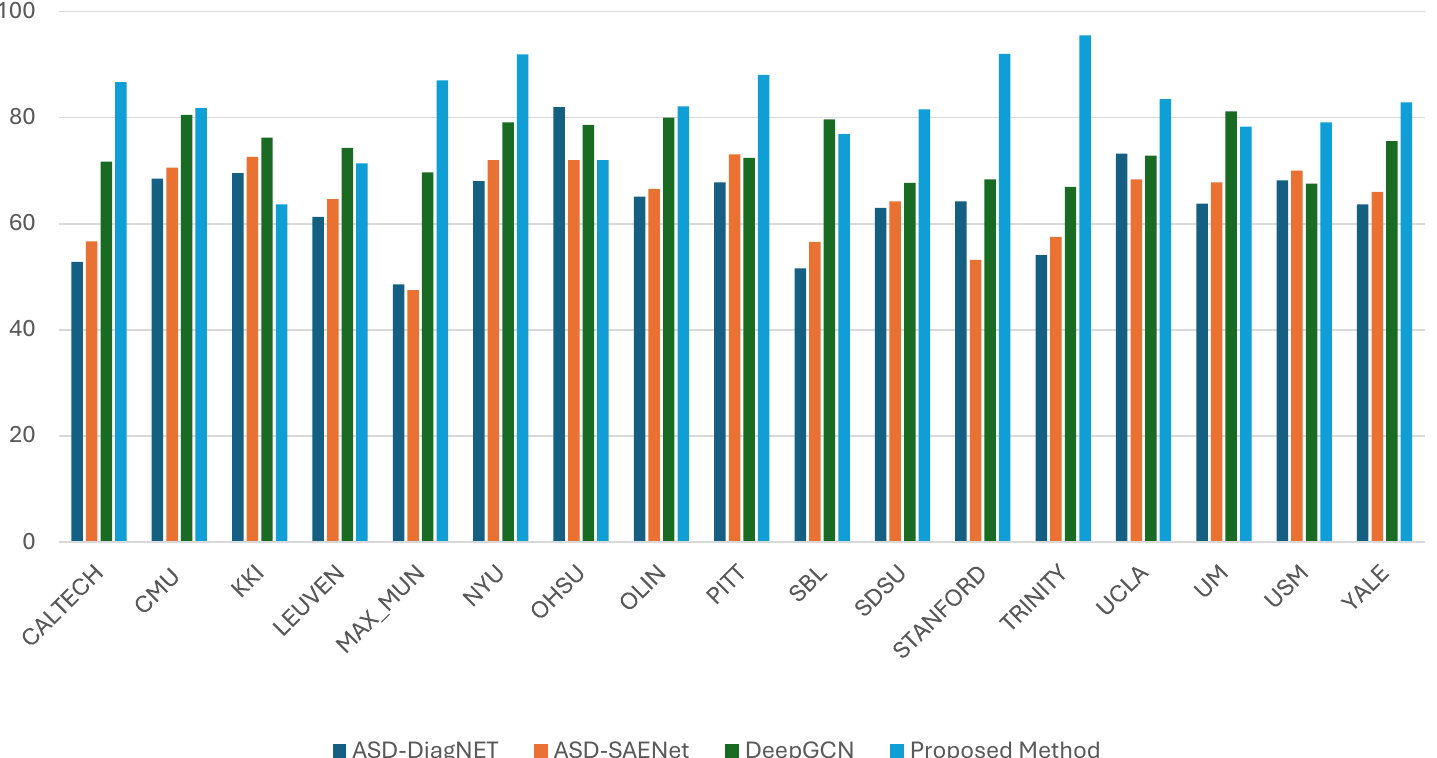}
    \caption{Bar plot comparing the performance of the proposed model with SOTA methods under LOSO cross-validation}
    \label{Bar Plots}
\end{figure}

\section{Discussion}

The experimental findings demonstrate that the proposed multimodal graph-based fusion framework effectively captures complementary functional and structural brain patterns for ASD classification. In particular, the integration of adaptive population graph modeling with asymmetric cross-attention-based multimodal fusion improved performance compared to single-modality and existing state-of-the-art approaches, as shown in Tables ~\ref{tab:ablation_results} and ~\ref{tab:sota_results}, respectively. These results suggest that jointly modeling inter-subject relationships and modality-specific representations provides a more discriminative and biologically meaningful characterization of brain connectivity alterations.

The ablation analysis indicates that rs-fMRI provides stronger discriminative power than sMRI when used independently, which is consistent with prior literature highlighting the sensitivity of functional connectivity to atypical network synchronization in ASD. Transformer-based and graph-based models operating on rs-fMRI, such as Brain Network Transformer \citep{kan2022brain}, Com-BrainTF \citep{bannadabhavi2023community}, and ASDFormer \citep{izadi2025asdformer}, have demonstrated that modeling long-range dependencies enhances classification performance. However, these approaches are primarily constrained to functional data. The results of the present study suggest that although rs-fMRI alone is competitive, incorporating structural connectivity further enriches the representation space by capturing anatomical organization that functional measures alone cannot fully characterize.

For sMRI combined with non-imaging data, the proposed framework substantially outperforms previously reported model FCN \citep{dong2025framework}. This improvement highlights the importance of population-graph modeling when structural connectivity is considered. Unlike conventional deep learning models that operate at the subject level, the adaptive graph construction enables non-imaging phenotypic attributes to guide inter-subject relationships, providing contextual information that enhances discrimination even when the imaging modality itself is less sensitive than rs-fMRI. These findings reinforce the relevance of incorporating demographic and acquisition-related factors into graph-based neuroimaging frameworks.

The superiority of the multimodal configuration further confirms that functional and structural connectivity encode complementary aspects of brain organization. While functional connectivity reflects dynamic synchronization patterns, structural connectivity represents relatively stable anatomical pathways. The asymmetric cross-attention mechanism enables selective information exchange between modalities, allowing functional embeddings to attend to structural representations in a context-aware manner. Compared to simple concatenation based fusion \citep{ashrafi2025enhanced}, which treats modality-specific embeddings independently before classification, cross-attention explicitly models inter-modality dependencies, leading to more expressive fused representations. This observation aligns with emerging evidence that attention-based fusion strategies provide advantages over static fusion schemes in multimodal neuroimaging analysis.

In comparison with state-of-the-art methods, the proposed framework achieves improved performance across both transformer and graph-based approaches. Population-graph models such as EV-GCN \citep{huang2020edge} demonstrated the effectiveness of jointly learning node embeddings and adaptive graph structures from rs-fMRI and non-imaging phenotypic data. Building upon this foundation, the present study extends the modeling capacity by incorporating structural connectivity and introducing asymmetric cross-attention for multimodal integration. Similarly, multimodal graph frameworks such as those reported in \citep{dong2025framework} and enhanced EV-GCN variants \citep{ashrafi2025enhanced} improve representation learning through multi-source information; however, they rely on simpler fusion strategies. The consistent gains observed in the current experiments suggest that explicitly modeling modality interaction, in addition to adaptive graph learning, contributes to improved discriminative performance.

The LOSO-CV evaluation provides further insight into the generalization capability of the framework under realistic multi-center conditions. Compared with ASD-DiagNet \citep{eslami2019asd}, SAENet \citep{almuqhim2021asd}, and DeepGCN \citep{wang2023multimodal}, the proposed model achieves the highest average accuracy and demonstrates superior performance across the majority of acquisition sites. This indicates that adaptive phenotypic-driven edge construction, combined with multimodal feature integration, enhances robustness to inter-site variability. The improved performance across both large cohorts and several smaller sites suggests that the model effectively leverages heterogeneous training data to learn more stable population-level representations.

Nevertheless, variability across certain smaller sites remains observable, particularly in cohorts with limited sample sizes. Such fluctuations are consistent with previously reported LOSO findings on ABIDE, where site-specific distribution shifts and demographic imbalance influence generalization performance. Although the proposed framework improves the overall average accuracy, these observations highlight the ongoing challenge of domain heterogeneity in multi-site neuroimaging studies.

\section{Conclusion}

In this study, we proposed a multimodal population-graph framework for ASD classification that integrates rs-fMRI, sMRI, and non-imaging phenotypic information within a unified learning strategy. By modeling subjects as nodes in an adaptive graph and learning inter-subject relationships through non-imaging phenotypic affinity, the framework effectively captures both individual connectivity patterns and population-level structure.

The introduction of an asymmetric cross-attention fusion mechanism allows functional representations to selectively incorporate complementary structural information, leading to improved discriminative performance compared to single imaging modality and simple fusion approaches. Experimental results under both 10-fold and LOSO cross-validation demonstrate the robustness and generalization of the proposed model across multi-site data.

Overall, this work emphasizes how crucial it is to create fusion strategies that use complementary information while respecting modality hierarchy. The proposed framework provides a scalable and interpretable approach for multimodal neuroimaging analysis and may serve as a foundation for future research in computer-aided diagnosis of neurodevelopmental and psychiatric disorders.

\section*{Acknowledgements}

This study was funded by Gesellschaft für Forschungsförderung Niederösterreich m.b.H. under project number: FTI24-G-024

\bibliographystyle{elsarticle-num} 
\bibliography{References}






\end{document}